\documentclass{article}

\PassOptionsToPackage{numbers}{natbib}

\usepackage[final]{nips_2017_nofoot}

\usepackage[utf8]{inputenc} 
\usepackage[T1]{fontenc}    
\usepackage{hyperref}       
\usepackage{url}            
\usepackage{booktabs}       
\usepackage{amsfonts}       
\usepackage{nicefrac}       
\usepackage{microtype}      
\usepackage{graphicx}
\usepackage{subfigure}
\usepackage{amssymb,amsmath,bm}

\title{Uncovering Latent Style Factors for Expressive Speech Synthesis}

\author{
  Yuxuan Wang, RJ Skerry-Ryan, Ying Xiao, Daisy Stanton,\\
\textbf{Joel Shor, Eric Battenberg, Rob Clark, Rif A. Saurous}\\  
      Google Research\\ Mountain View, CA \\
  \texttt{\{yxwang,rjryan,rif\}@google.com}
}

\begin{document}

\maketitle

\newcommand\blfootnote[1]{%
  \begingroup
  \renewcommand\thefootnote{}\footnote{#1}%
  \addtocounter{footnote}{-1}%
  \endgroup
}

\begin{abstract}
Prosodic modeling is a core problem in speech synthesis. The key challenge is producing desirable prosody from textual input containing only phonetic information. In this preliminary study, we introduce the concept of ``style tokens'' in Tacotron, a recently proposed end-to-end neural speech synthesis model. Using style tokens, we aim to extract independent prosodic styles from training data. We show that without annotation data or an explicit supervision signal, our approach can automatically learn a variety of prosodic variations in a purely data-driven way. Importantly, each style token corresponds to a fixed style factor regardless of the given text sequence. As a result, we can control the prosodic style of synthetic speech in a somewhat predictable and globally consistent way.

\end{abstract}

\section{Introduction}
A text-to-speech (TTS) model generates speech from textual input. Ideally, the generated speech should convey the correct message (intelligibility) while sounding like human speech (naturalness) with the right prosody (expressiveness). Most speech synthesis systems are focused on solving the first two issues. For example, traditional concatenative and parametric synthesis models can deliver intelligible speech with certain degrees of naturalness \cite{wan2017google}. Recent advances in neural modeling of speech significantly improve the naturalness of synthetic speech. For example, Tacotron \cite{wang2017tacotron} can produce very good flow and rhythm directly from grapheme inputs. WaveNet \cite{arik2017deep, van2016wavenet} substantially improves synthetic audio fidelity using linguistic and prosodic features as input.  Most existing synthesis models, however, do not explicitly model prosody. Despite having important applications such as conversational assistants and long-form reading, expressive TTS is still considered an important open problem.

In most scenarios, TTS models are only given text inputs at runtime, with no acoustic reference. Prosodic variation is inherently multi-scale. Local changes in pitch and speaking duration can convey semantic meaning while global properties such as overall pitch trajectory can convey mood and emotion. If the training data have high prosodic variation, synthesis becomes challenging due to the difficulty of solving the underlying inverse problem. Even when the training data contain mostly neutral prosody, nuanced prosodic variations still occur at a fine time resolution. In this work, we introduce the concept of \textit{``style tokens''}, which can be viewed as latent variables capturing prosodic variations that text input alone cannot capture. Style tokens can be learned in an unsupervised fashion, and do not require annotated labels, which can be both noisy and expensive to obtain.  We implement style tokens in the Tacotron model and demonstrate that they indeed correspond to distinct prosodic style factors, enabling some degree of prosody control by specifying the desired style in inference.


\begin{figure*}[t]
\centering
\includegraphics[scale=0.68]{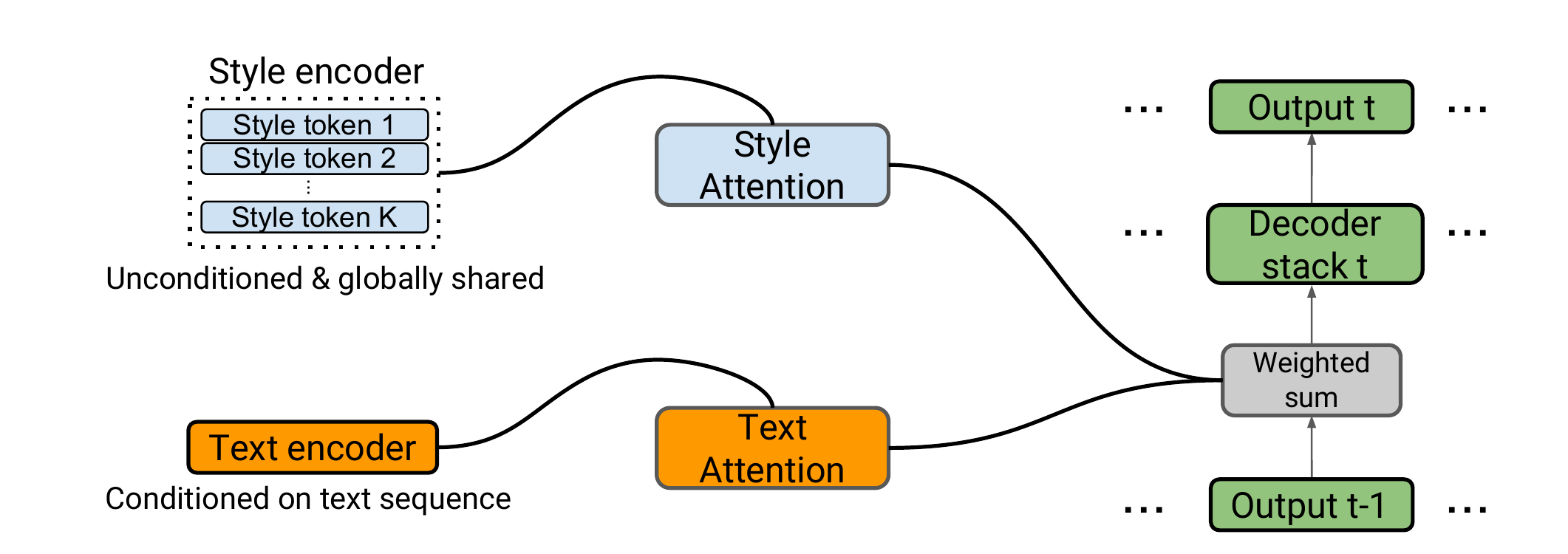}
\caption{{Model architecture cartoon based on Tacotron \cite{wang2017tacotron}. To learn style tokens in Tacotron, we add an additional style encoder and the corresponding attention pathway. See \cite{wang2017tacotron} for Tacotron architecture details.}}
\label{fig:model}
\end{figure*}

\section{Model Architecture}
Our model is based on Tacotron \cite{wang2017tacotron}, a recently proposed end-to-end speech synthesis model that predicts linear spectrograms directly from character (or phoneme) sequences, and that uses Griffin-Lim or a neural vocoder to convert spectrograms to waveforms. Tacotron is based on the sequence-to-sequence with attention framework \cite{bahdanau2014neural}, which consists of an encoder of the text sequence, a decoder that predicts mel spectrograms, an attention module that learns to align text with acoustics, and an optional post-processing net that predicts linear spectrograms.


Ideally, an expressive TTS model should allow explicitly-controlled prosody selection during synthesis. Since Tacotron takes only text as input, in order to reconstruct training signal accurately, it must learn to store any prosodic information implicitly in its weights, which we don't have explicit control over. To allow explicit prosodic control, we introduce a dedicated network component to Tacotron, augmenting the existing text-encoder attention pathway with a new style attention pathway, as depicted in Figure \ref{fig:model}. The new attention module attends to a style encoder, which takes $K$ ``style tokens'' as input, and outputs embedding vectors for the style attention to attend to. In the decoder, we compute the text attention and style token attention heads in parallel, and combine their context vectors via a weighted sum. A ``controller'' layer is tasked with computing the weighting of the two context vectors. The input to the controller is the same input used to compute the attention context vectors (i.e. teacher forced signal or the previous decoder step output). The rest of the Tacotron model remains the same.


Style tokens can be interpreted as latent variables (non-probabilistic in this work but can be extended to be probabilistic), encoding variations that the text input alone can not capture. Their embedding values are randomly initialized and automatically learned by backpropagation, where the learning is guided only by the reconstruction loss on the decoder. Therefore, the learning of style tokens per se is entirely unsupervised. While it's possible to produce multiple prosodies by other methods (such as priming the decoder initial states, as done in \cite{taigman2017voice}), using attention-based style tokens has several benefits. First, attention helps learn a decomposition of the overall prosodic style, encouraging interpretable tokens with independent prosodic styles. This is similar to learning a dictionary of style atoms that can be combined to reproduce the overall style. Furthermore, an attention mechanism learns a combination of style tokens at the decoder's time resolution, which enables time-varying prosody manipulation.


Finally, we point out an important difference between the style encoder and the text encoder. While the text encoder is conditioned on an input text sequence, the style encoder takes no inputs, and the tokens are \emph{shared} across all training sequences. In other words, the style encoder computes priors for the training set, whereas the text encoder computes posteriors (conditioned on individual input sequences). This design allows style tokens to capture text-independent prosodic variation, which enables controllability in inference.

\begin{figure}[t]
\centering
\centerline{
\subfigure[]{
\label{fig:f0:a}
\includegraphics[scale=0.4]{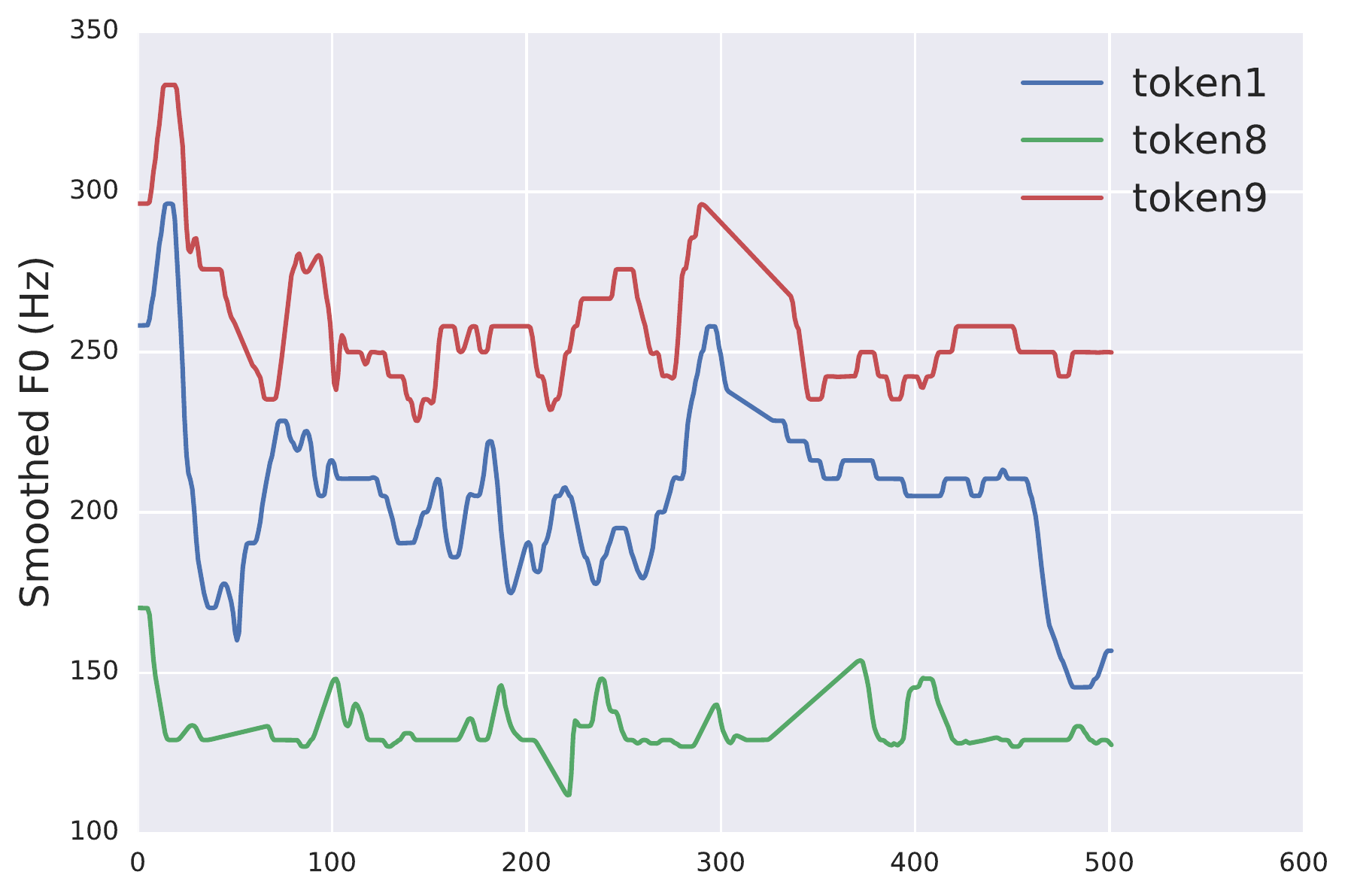}
}
\subfigure[]{
\label{fig:f0:b}
\includegraphics[scale=0.4]{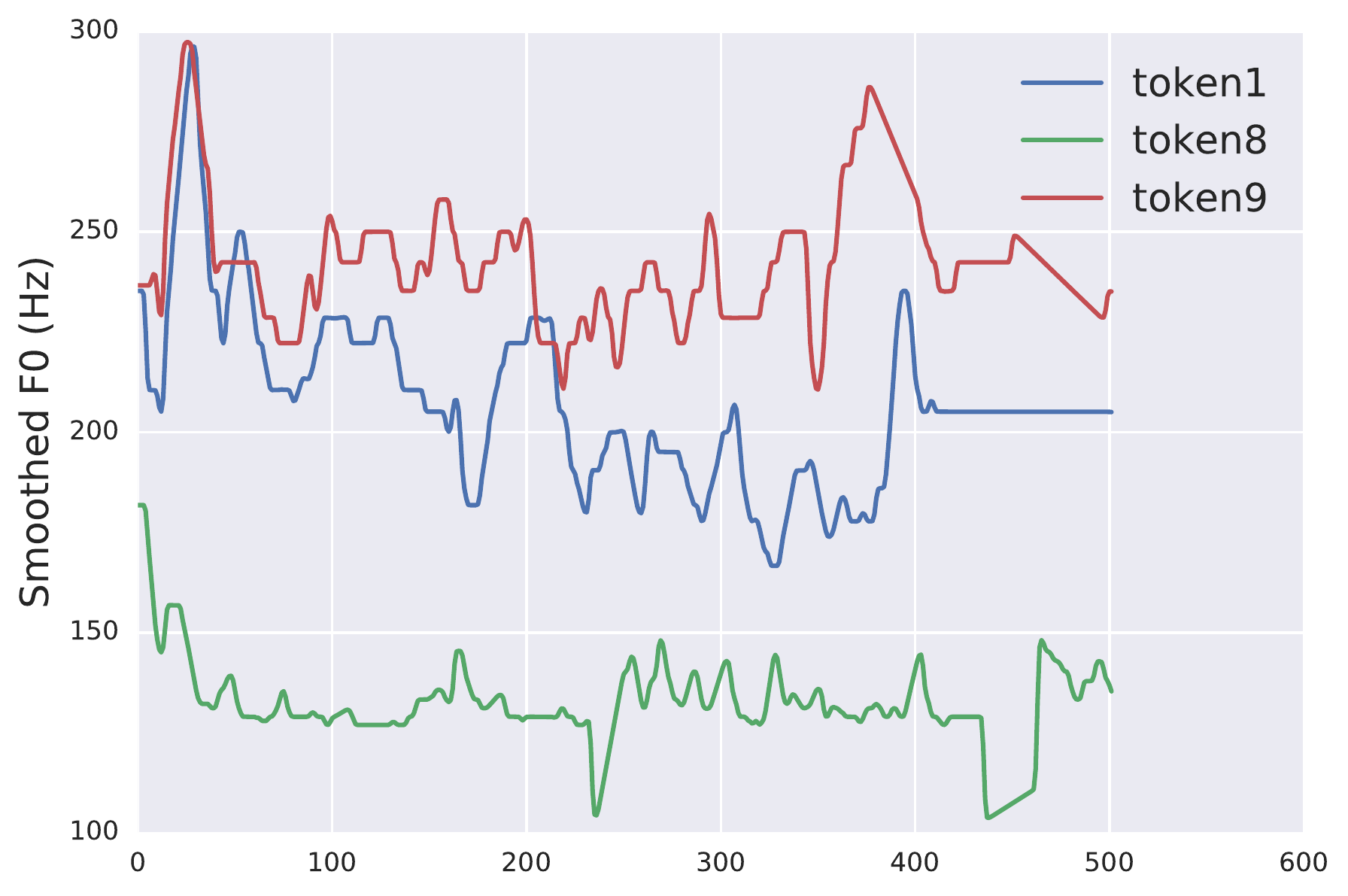}
}
}
\caption{Smoothed F0 trajectories of two different utterances (left and right), synthesized using three style tokens. See text for details.}
\label{fig:f0}
\end{figure}

\section{Related Work}
\label{sec:related}
Prosody and speaking style modeling have been studied since the era of HMM-based TTS research. For example, \cite{eyben2012unsupervised} proposes a system that first clusters the training set, and then performs HMM-based cluster-adaptive training. \cite{nose2007style} proposes estimating the transformation matrix for a set of predefined style vectors.

While sharing many ideas with HMM-based TTS, deep-neural-network-based TTS enables easy and flexible conditioning of external attributes, such as one-hot speaker codes \cite{arik2017deep, luong2017adapting}. These models can capture overall speaker characteristics, but most of them can't model or control specific speaking styles. This argument applies similarly to other conditioning attributes, such as emotion \cite{luong2017adapting} and prominence \cite{malisz2017controlling}. More generally, these conditioning inputs are termed as control vectors in TTS literature \cite{watts2015sentence, henter2017principles}. In particular, \cite{henter2017principles} provides a latent variable interpretation of control vectors, which is similar to this work. However, in addition to model differences such as style attention, \cite{henter2017principles} aims to learn nuances within a predefined set of speech emotion classes, whereas this work does not rely on a predefined partition and is more general.

In summary, our method differs from existing work in the following aspects. First, by designing style tokens on top of the Tacotron architecture, we demonstrate prosody and style control as part of an end-to-end speech system. In constrast to traditional TTS systems, which only represent prosody and style in an acoustic model, our approach allows these latent factors to be captured throughout the model. Second, style tokens are computed from a differentiable attention mechanism, which helps learn a set of mutually independent prosodic features that can then be flexibly recombined as the user desires. Finally, our method is unsupervised; it does not require global (e.g. emotion) or local (e.g. prominence) annotations, which can be unreliable and expensive to obtain. If these are available, however, style tokens can be easily adapted to incorporate them, and can also easily work with other explicit control vectors.

\let\oldsim\sim 
\renewcommand{\sim}{{\oldsim}}
\section{Results}
We train a style token model based on an improved version of Tacotron \cite{wang2017tacotron} (details omitted), which achieves $\sim$4.0 mean opinion score on the eval set used in \cite{wang2017tacotron}. The model uses 10 style tokens with content-based RNN attention. The weights used to mix the text and style attention are predicted by a single layer MLP with sigmoid outputs. We use a similar single-speaker corpus as in \cite{wang2017tacotron}; while the speaker primarily speaks in a neutral prosody, a small subset of the corpus uses more expression (including performing as a game show host, reading jokes and poems, etc). We aim for our model to capture these variations, even though they appear in a minority of the training corpus.

After training, we can ``listen'' to each style token~\footnote{Sound demos can be found at \url{https://google.github.io/tacotron/publications/uncovering_latent_style_factors_for_expressive_speech_synthesis/}} by forcing the style attention to attend to the specified token. Since the training process learns to rely on a mixture of tokens, this method can lead to unintelligible speech, but it nonetheless provides a rough idea of the prosodic style each token corresponds to. To synthesize speech in a specific style, we can broadcast-add the embedding vector of the selected style token to the full style embedding matrix, thereby biasing synthesis towards the specified style. Similarly, we can mix different styles by consecutively broadcast-adding or linearly interpolating style embedding vectors. Other mixing methods can achieve similar effects, and we defer a study of them to future work.

To quantify the behavior of style tokens, we generate smoothed F0 trajectories from synthetic speech mixed with three different style tokens. As can be heard from the audio demos, ``token 1'' roughly corresponds to a sloppy style with a normal pitch range, ``token 8'' roughly corresponds to a robotic-sounding style, and ``token 9'' roughly corresponds to high-pitched voice. As shown in Figure \ref{fig:f0:a}, these styles are somewhat reflected in the smoothed F0 trajectories. E.g., ``token 9'' tends to have higher pitch than the other two, and the pitch track of ``token 8'' stays flat and low. The same trend can be seen from Figure \ref{fig:f0:b}, which is produced from a different utterance, demonstrating that style tokens operate independently of text input.

\begin{figure}[t]
\centering
\centerline{
\subfigure[]{
\label{fig:controller:a}
\includegraphics[scale=0.4]{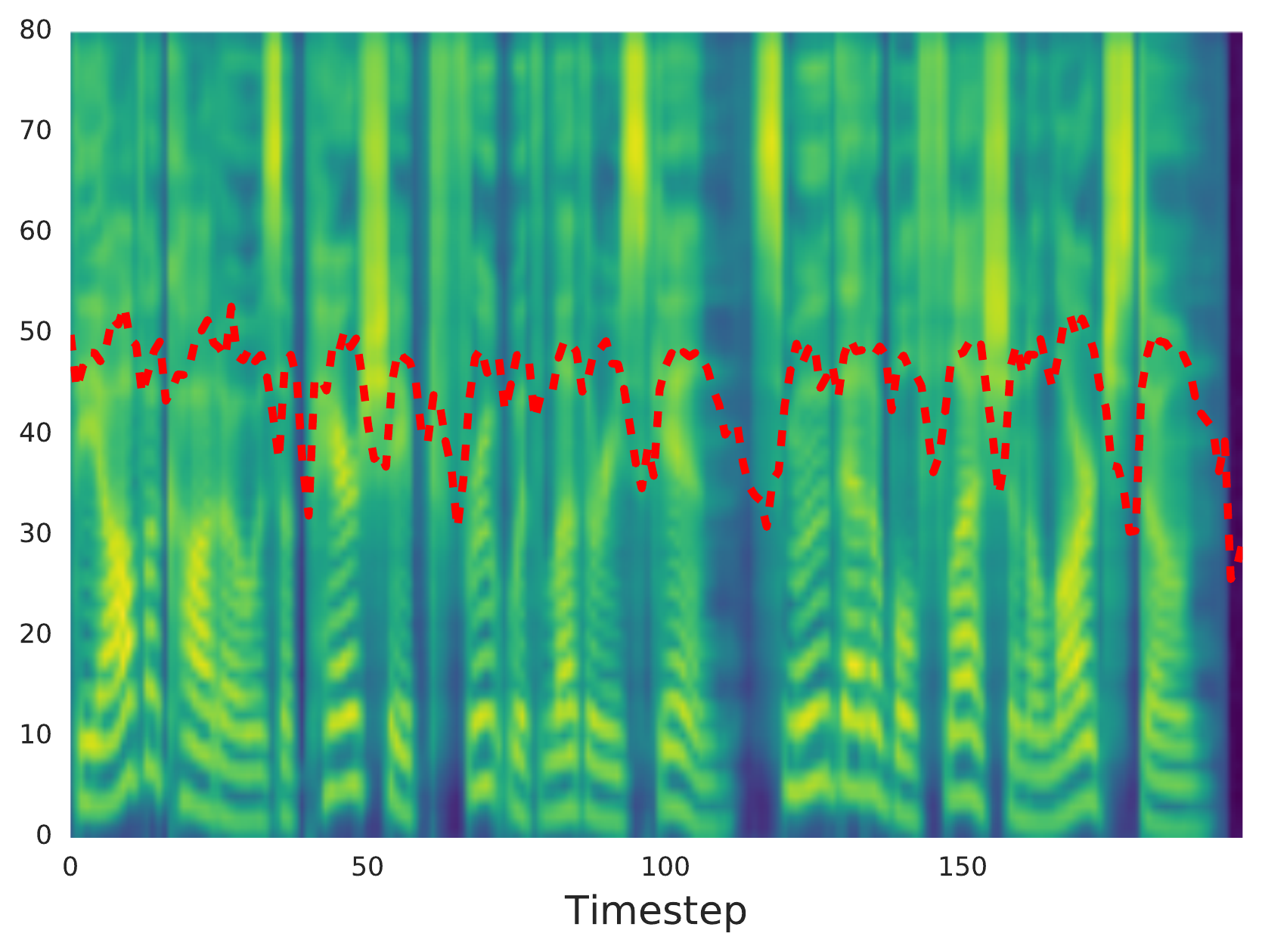}
}
\subfigure[]{
\label{fig:controller:with}
\includegraphics[scale=0.4]{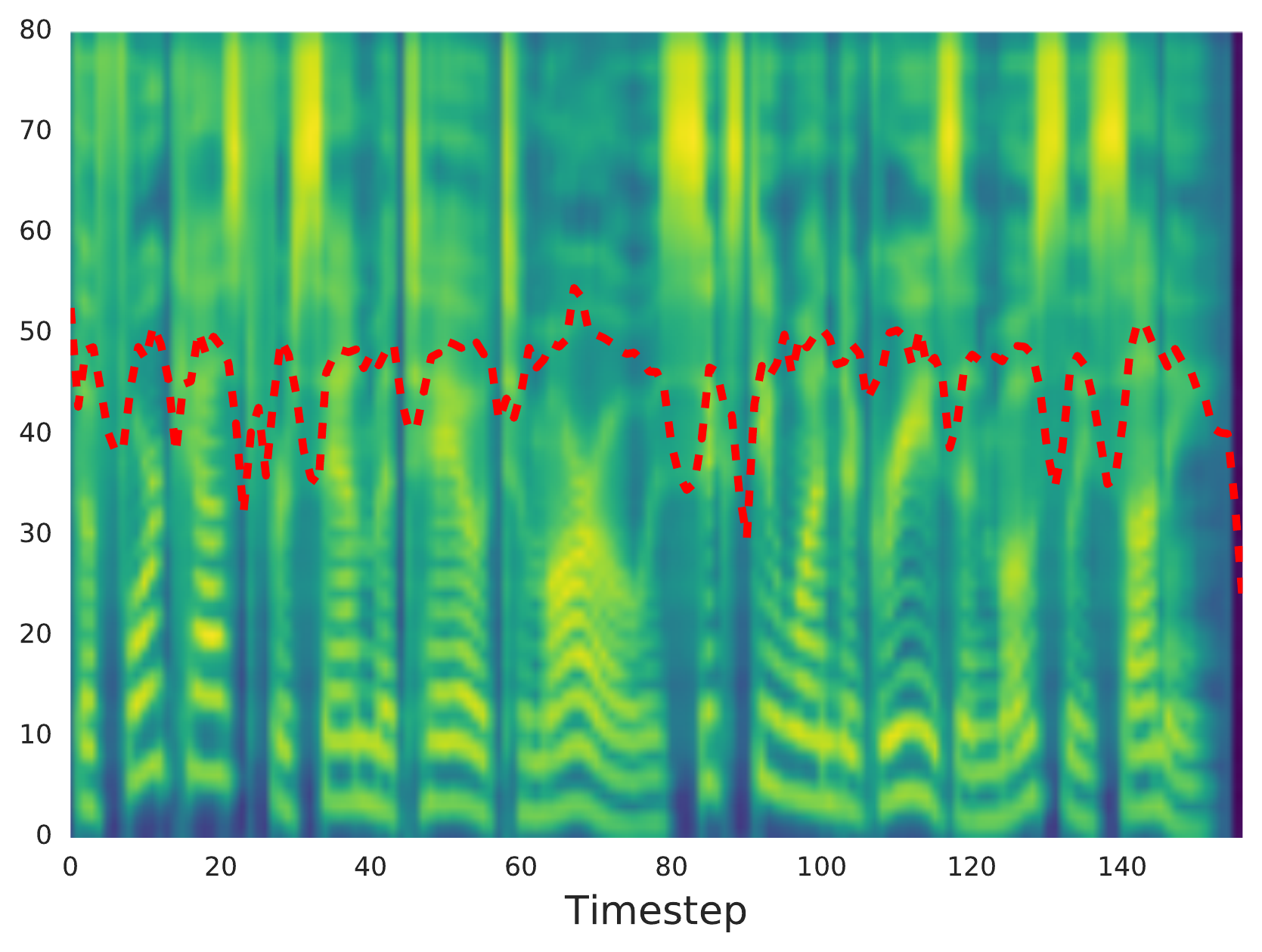}
}
}
\caption{{Visualizations of two different utterances (left and right), showing predicted mixing weights (dashed red lines) for the text attention overlaid on top of the predicted mel spectrograms. See text for details.}}
\label{fig:controller}
\end{figure}

Figure \ref{fig:controller} shows the predicted mixing weights for the text attention (in $[0,1]$, linearly re-scaled for visualization), which are overlaid on top of the predicted mel spectrograms. It is interesting to note that the peaks and valleys of the weights align well with speech segmental boundaries. We hypothesize that the model alternates between determining content (based on text tokens) and rendering its style (based on style tokens).

\section{Discussions and Future Work}
We have proposed a simple new method, which we call style tokens, that can uncover latent style factors for prosody modeling and control. The style token idea fits very well with end-to-end models such as Tacotron, and can be implemented by adding a style attention pathway in addition to the text attention pathway. Learning style tokens is unsupervised and does not require labelled data.

We note that this work is highly preliminary and there is significant room for improvement and expansion. In particular, we can treat the style encoder as an external memory to leverage advances from memory-augmented neural nets \cite{graves2014neural}. Future work will explore the combination of style tokens and explicit control signals on more expressive datasets.

\bibliographystyle{IEEEtran}
\bibliography{uncover_bib}

\end{document}